\documentclass{article} 
\usepackage{iclr2026_conference,times}

\usepackage{amsmath,amsfonts,bm}

\def\Secref#1{Section~\ref{#1}}

\def\eqref#1{equation~\ref{#1}}

\def\1{\bm{1}}

\DeclareMathAlphabet{\mathsfit}{\encodingdefault}{\sfdefault}{m}{sl}
\SetMathAlphabet{\mathsfit}{bold}{\encodingdefault}{\sfdefault}{bx}{n}

\usepackage{amssymb}
\usepackage{algorithm}
\usepackage{algpseudocode}
\usepackage[dvipsnames,table]{xcolor}
\usepackage{graphicx}
\usepackage{booktabs}
\usepackage{multirow}
\usepackage{wrapfig}
\usepackage[colorlinks=true]{hyperref}
\usepackage{url}

\hypersetup{
    citecolor=YellowGreen,
}

\definecolor{catgray}{gray}{0.9}

\title{Ms. Forcing: Efficient Streaming Video Generation with Multi-Scale Patchification and Attention}

\author{
\textbf{Zekun Li}$^{1*}$ \quad
\textbf{Xiaoyan Cong}$^{1*}$ \quad
\textbf{Hongyu Li}$^{1}$ \quad
\textbf{Zhiyang Dou}$^{2}$
\\[0.35em]
\textbf{Chuan Guo}$^{3}$ \quad
\textbf{Abhay Mittal}$^{3\dagger}$ \quad
\textbf{Sizhe An}$^{3\dagger}$ \quad
\textbf{Srinath Sridhar}$^{1\ddagger}$
\\[0.6em]
$^{1}$Brown University \quad
$^{2}$Massachusetts Institute of Technology \quad
$^{3}$Meta
}

\iclrfinalcopy 
\begin{document}

\maketitle
\fancyhead{} 
\begingroup
\renewcommand{\thefootnote}{\fnsymbol{footnote}}
\footnotetext[1]{Equal contribution. \quad
$^{\dagger}$Joint project leads. \quad
$^{\ddagger}$Corresponding author.}
\endgroup

\begin{figure}[ht]
    \centering
    \includegraphics[width=\linewidth]{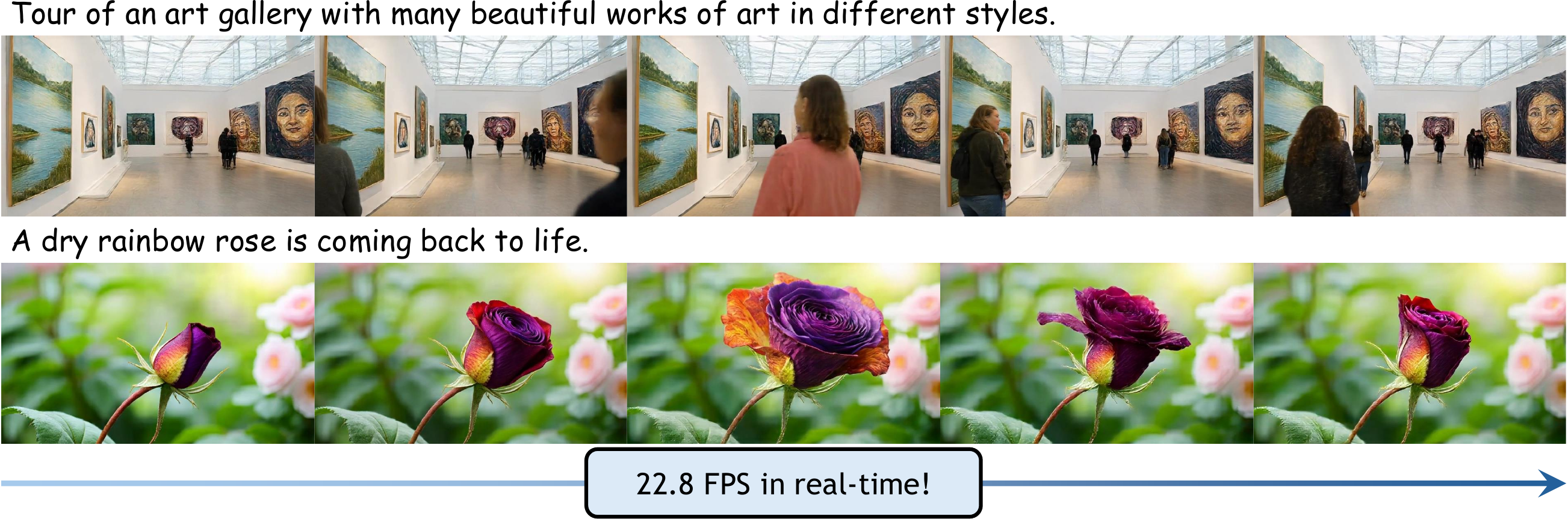}
    \caption{\textbf{Ms.~Forcing} is a real-time streaming video generation framework ($22.8$ FPS) built on rolling diffusion.
    It combines our novel Multi-Scale Patchification (MSP) and Multi-Scale Self-Attention (MSSA), which adapt the spatial granularity of computation to each state's noise level in the rolling denoising window, with Homogeneous-Noise-Level DMD (H-DMD), which reduces the mismatch between DMD training sequences and inference-time rollouts.
    }
    \label{fig:teaser}
\end{figure}

\begin{abstract}
Streaming video diffusion models have made substantial progress toward interactive and dynamic world simulation, but the nested autoregressive and denoising loops of conventional next-frame generation hinder real-time deployment.
Recent rolling-window methods pipeline denoising across multiple consecutive frames at different noise levels, improving throughput and long-horizon stability.
However, they tokenize every state at the same fine spatial granularity, leaving substantial noise-dependent redundancy in the joint denoising window.
We propose \textbf{Ms.~Forcing}, an efficient streaming video generation paradigm that adapts spatial granularity to each state's noise level.
Its \emph{Multi-Scale Patchification} (MSP) assigns coarser patches to noisier states, reducing the active-window token count by 45\%, while \emph{Multi-Scale Self-Attention} (MSSA) matches the density of visible non-sink keys and values to each query scale to further reduce attention cost.
Because both schedules are fixed by window position, Ms.~Forcing retains a static, hardware-friendly computation graph.
We further introduce \emph{Homogeneous-Noise-Level DMD} (H-DMD), which assembles each fake video from clean predictions sharing a same source noise level, thereby reducing the mismatch between DMD training sequences and inference-time rollouts.
The multi-scale design helps offset the additional training cost of backpropagating through overlapping windows.
We include both quantitative and qualitative experiments to show that Ms.~Forcing reaches \textbf{22.84 FPS on a single H200 GPU}, 39.6\% faster than Rolling Forcing, while significantly improving VBench scores in both short video and long video generation setting.
\end{abstract}
\section{Introduction}
Recent large-scale video diffusion models~\citep{wan2025wan,hacohen2026ltx,team2025longcat} generate realistic videos with intricate dynamics and high visual quality.
However, synthesizing a complete clip through iterative bidirectional denoising incurs high latency and precludes on-demand output.
Interactive applications such as generative world models~\citep{bruce2024genie,gao2026infinite} and real-time virtual avatars~\citep{sun2026streamavatar,huang2026video} instead require \emph{streaming} generation: frames must be emitted sequentially with low latency while remaining coherent over long horizons.
Delivering real-time throughput without sacrificing visual fidelity, motion quality, or long-horizon stability remains a central challenge.

Recent methods enable streaming by distilling slow bidirectional diffusion models into few-step causal autoregressive generators~\citep{yin2025slow,huang2026self}.
A conventional causal sampler nests an inner denoising loop for each new frame inside an outer autoregressive loop.
Each frame is finalized before the next is generated, so errors propagate along the autoregressive trajectory.
Also, inference efficiency constraints will limit the history context length and further exacerbate long-horizon drift~\citep{huang2026self}.
\citet{liu2026rolling} instead introduce Rolling Forcing (RF), which pipelines $T$ denoising steps along the temporal axis through a joint window of $T$ consecutive frames at progressively increasing noise levels.
Each pass advances all window states by one noise level and emits the clean head frame.
Bidirectional attention within the window lets frames refine one another before emission, expanding the denoising context and reducing error accumulation while preserving causal output.

RF trades nested sampling for a heavier forward pass: all $T$ active states are tokenized at the backbone's native patch size despite their different noise levels.
Pruning or compressing history KVs~\citep{guo2026efficient,yang2026longlive} reduces the cost of past context but does not address this mixed-noise active window.
Diffusion denoising is inherently coarse-to-fine, i.e., high-noise states recover low-frequence structure, whereas low-noise states refine high-frequence details~\citep{kim2026ddit,jin2025pyramidal}.
In RF, each state's noise level is fixed \emph{a priori} by its window position, enabling a static multi-scale schedule without input-dependent routing.
Motivated by the zero-shot probes of redundancy in the joint denoising window in \Secref{sec:motivation}, we propose \emph{Multi-Scale Patchification} (MSP), which assigns coarser patchification to noisier states, and \emph{Multi-Scale Self-Attention} (MSSA), which matches the density of visible non-sink KVs to each query's patch granularity while preserving the full-resolution sink frame.
Under our 5-step RF schedule, MSP reduces tokens in denoising window by $45\%$, and MSSA further removes $12.9\%$ of post-MSP dot-product attention computation.
Both are fixed by window position, yielding a static compute graph.

The reduced training cost also enables a more faithful construction for Distribution Matching Distillation (DMD; \citealp{yin2024improved,yin2024one}).
To control memory, RF builds its fake video from non-overlapping windows by concatenating all $T$ predicted clean states from each selected window.
Adjacent frames therefore originate from different rollout noise levels, whereas inference concatenates head predictions from consecutive windows.
This heterogeneity creates a mismatch between DMD training sequences and inference-time emitted rollouts.
We propose \emph{Homogeneous-noise-level DMD} (H-DMD) for joint denoising, which samples one window position per update and gathers its predicted clean state from every consecutive window.
All frames in the assembled fake video then share one source noise level, yielding a consistent temporal marginal without changing the DMD objective.
Although it requires gradients through overlapping windows, MSP and MSSA make it tractable and remove the need for auxiliary Self-Forcing~\citep{huang2026self} rollout updates.

Together, these designs form \textbf{Ms.~Forcing}.
At $832\times480$ resolution, it reaches $22.84$ FPS on a single H200 GPU ($39.6\%$ faster than RF), improves five-second VBench quality and semantic scores over RF by $0.29$ and $1.25$, and reduces 60-second quality drift from $2.227$ to $1.700$ while maintaining competitive quality.
Our contributions are:
\begin{itemize}
    \item We introduce \textbf{Multi-Scale Patchification} and \textbf{Multi-Scale Self-Attention}, a static noise-matched schedule that reduces tokens in denoising window by $45\%$ and post-MSP attention computation by a further $12.9\%$.
    \item We identify a training--inference mismatch introduced by RF's heterogeneous DMD assembly and propose \textbf{Homogeneous-Noise-Level DMD}, which improves distillation effectiveness and generation quality without requiring auxiliary Self-Forcing training.
\end{itemize}

\section{Related Work}
\paragraph{Autoregressive Video Diffusion Models.}
Large-scale video diffusion models (VDMs)~\citep{wan2025wan,hacohen2026ltx,team2025longcat,yang2025cogvideox,kong2024hunyuanvideo,seedance2026seedance,cong2026viva,du2026videogpa,li2026genhsi} typically synthesize entire clips using bidirectional attention, making them unsuitable for low-latency streaming.
To transform these slow bidirectional VDMs into efficient streaming generators, recent work combines autoregressive modeling with iterative diffusion denoising under several formulations~\citep{chen2024diffusion}.
Most existing streaming video generators~\citep{yin2025slow,huang2026self,lu2026reward} adopt a nested sampling process: an outer autoregressive loop generates frames sequentially, while an inner diffusion loop progressively denoises each new frame.
A complementary line of work~\citep{ruhe2024rolling,kim2024fifo,sun2025ar,liu2026rolling} pipelines the denoising steps along the temporal axis, jointly updating consecutive blocks at progressively increasing noise levels within a rolling window and emitting one clean block per forward pass.
To mitigate exposure bias and improve generation quality, recent approaches further apply distribution matching distillation (DMD) to self-generated rollouts, thereby better aligning training with the inference-time distribution~\citep{huang2026self}.

\paragraph{Efficient Video Diffusion.}
Beyond reducing the number of denoising steps through distillation, complementary acceleration methods eliminate redundant computation or lower the cost of each model evaluation.
For efficient and effective streaming inference, sliding-window attention and persistent sink frames bound the history context length while retaining critical long-range context~\citep{yang2026longlive,shin2026motionstream}.
Recent methods further reduce history cost through KV-cache token pruning or compression~\citep{guo2026efficient,chen2026pyramid,yi2025deep}.
Other acceleration strategies reuse intermediate features across adjacent denoising steps~\citep{liu2025reusing,liu2025timestep}, perform staged coarse-to-fine generation~\citep{jin2025pyramidal,ding2026surf,zheng2026multi}, or use adaptive patch schedules during denoising~\citep{kim2026ddit}.
Most of them rely on runtime importance estimation or adaptive routing, which can introduce input-dependent and irregular execution.
More importantly, acceleration within the joint denoising window~\citep{ruhe2024rolling,liu2026rolling} remains largely unexplored.
We exploit the deterministic correspondence between window position and noise level to assign noise-matched patch sizes and history receptive fields a priori, yielding a static, hardware-friendly schedule.

\section{Preliminary and Motivation}
\subsection{Preliminary: Autoregressive video diffusion models}
Autoregressive video diffusion factorizes an $N$-frame sequence $\boldsymbol{x}^{1:N}$ follows $p(\boldsymbol{x}^{1:N})=\prod_{i=1}^{N}p(\boldsymbol{x}^{i}\mid\boldsymbol{x}^{<i})$, where each $\boldsymbol{x}^{i}$ may represent a short frame chunk, referred to as a ``frame'' for simplicity.
For efficient streaming computation, recent works~\citep{lu2026reward,yang2026longlive} organize context tokens as $[\,\text{sink}\mid\text{recent}\mid\text{now}\,]$ under causal attention and generate each new block through few-step denoising~\citep{yin2025slow,huang2026self,yang2026longlive}.
Given a $T$-step diffusion ($t_{1}\le\cdots\le t_{T}$) model $p_{\theta}$, the frame $i$ generation can be formulated as
\begin{equation}
    p(\boldsymbol{x}^{i}_{t_{0:T}}\mid\boldsymbol{x}^{<i}) = \prod_{j=1}^{T} p_{\theta}(\boldsymbol{x}^{i}_{t_{j-1}}\mid \boldsymbol{x}^{i}_{t_{j}}, \boldsymbol{x}^{<i}) \,, \qquad \boldsymbol{x}^{i}_{t_{T}}\sim\mathcal{N}(0,\mathbf{I}) \,.
\end{equation}
While they enable consistent sequential generation, their strictly causal frame prediction causes each frame to inherit errors from its predecessors, allowing small imperfections to compound over long horizons and eventually leading to noticeable drift and quality degradation.
To this end, Rolling Forcing (RF) pipelines the $T$ denoising steps of each generated frame along the temporal axis via a joint rolling window containing $T$ consecutive frames assigned progressively higher noise levels $t_1\le\cdots\le t_T$.
Rather than fully denoising each frame before moving to the next, RF expands the denoising context length and allows frames to refine one another through bidirectional attention before finalization, reducing local errors and long-horizon drift while preserving causal streaming.
At each rolling, RF jointly denoises frames in the window by one noise level in a single forward pass, sampling from the transition
\begin{equation}
    p_{\theta}\!\left(\boldsymbol{x}^{i:i+T-1}_{t_{0:T-1}}\,\middle|\,\boldsymbol{x}^{i:i+T-1}_{t_{1:T}},\,\boldsymbol{x}^{<i}\right) \,,
    \qquad \boldsymbol{x}^{i+T-1}_{t_{T}}\sim\mathcal{N}(0,\mathbf{I}) \,.
\label{eq:roll_window}
\end{equation}
Position $l$ contains $\boldsymbol{x}^{i+l-1}_{t_l}$, ordered from the near-clean head at $t_1$ to the near-noise tail at $t_T$.
After each pass, RF emits the denoised head frame, advances the remaining frames toward $t_0$, and appends a fresh noise frame $\boldsymbol{x}^{i+T}_{t_T}\sim\mathcal{N}(0,\mathbf{I})$ at the tail, allowing the window to roll forward indefinitely.

\subsection{Zero-Shot Probes of Redundancy in the Joint Denoising Window}
\label{sec:motivation}
Diffusion denoising proceeds from coarse to fine: denoising at high noise levels primarily establishes global structure, whereas low-noise steps refine local details.
This observation has motivated prior diffusion models to process high-noise states at lower spatial resolutions~\citep{kim2026ddit,jin2025pyramidal}.
Unlike conventional denoising inference, which processes inputs at a single noise level during each denoising pass, RF jointly processes a mixed-noise-level input within the bidirectional denoising window.
Yet RF patchifies all frames in this joint denoising window at a uniform spatial granularity.
We therefore test whether this mixed-noise denoising window can instead be represented as a mixed-granularity input sequence based on the noise schedule.
To this end, we adopt two perturbations during pretrained RF rollouts in zero-shot, probing the sensitivity of the hybrid granularity of patchification and self-attention via frame-wise LPIPS~\citep{zhang2018unreasonable}.

\begin{wrapfigure}{r}{0.497\linewidth}
    \centering
    \vspace{-\intextsep}
    \includegraphics[width=\linewidth]{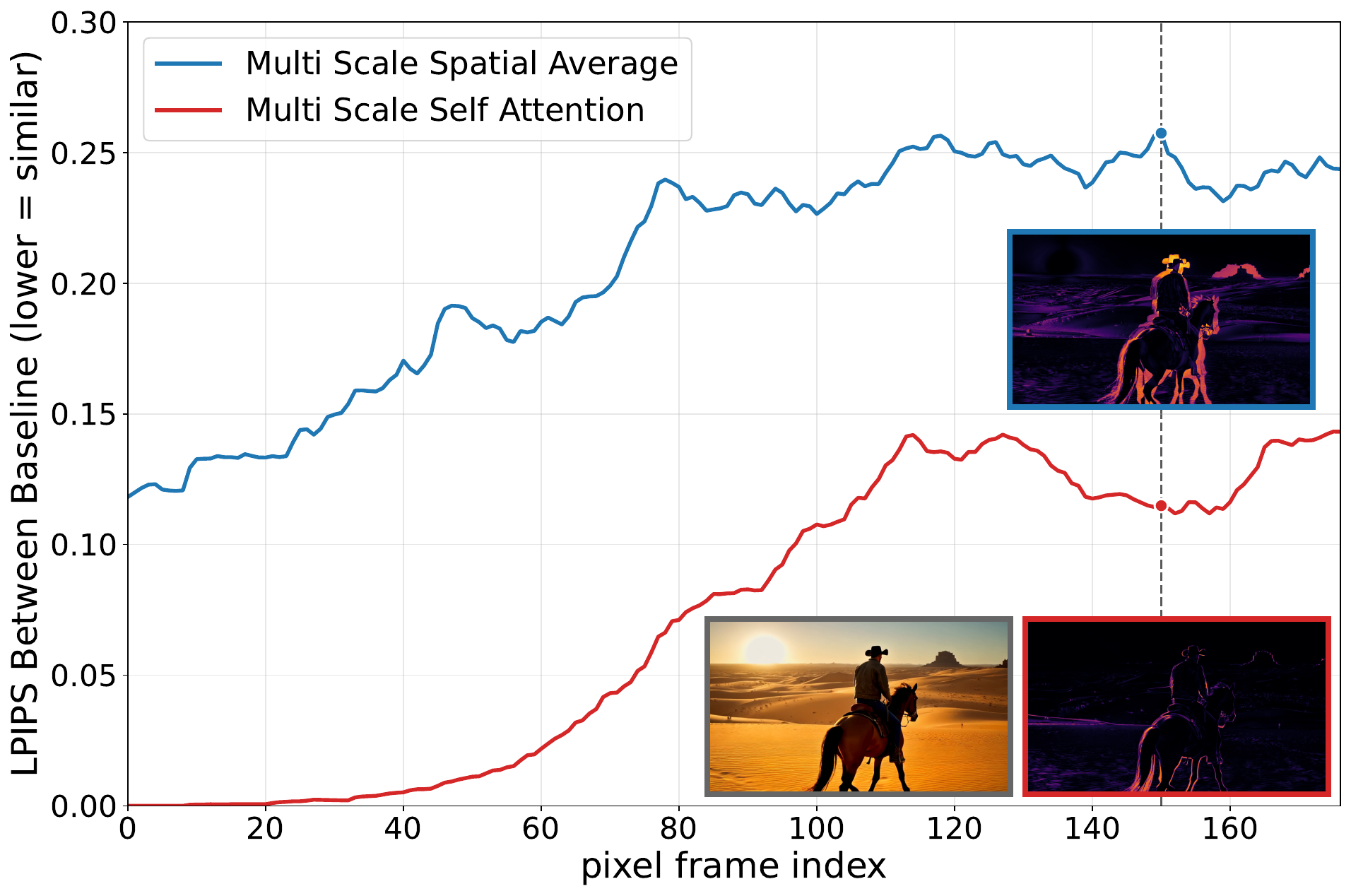}
    \vspace{-0.6cm}
    \caption{\textbf{Zero-shot probes of mixed granularity.}
    Frame-wise LPIPS ($\downarrow$) between each perturbed RF and the original rollout across $177$ pixel frames. 
    For $5$-step RF, we treat the three highest-noise levels as coarse. 
    Blue: After native patchification, we apply spatial average for frames at these levels;
    Red: For \emph{queries} at these levels, MSSA spatially subsamples their attention-visible \emph{keys} and \emph{values} with a stride of $2$.}
    \label{fig:motivation}
    \vspace{-0.7cm}
\end{wrapfigure}

\paragraph{Probing patchification granularity.}
Directly applying mixed-granularity patchification in zero-shot is infeasible because each new patch size requires learned patch-embedding and de-embedding layers.
We therefore emulate coarse patchification after native tokenization: at selected noisy window positions, we partition the spatial token grid into non-overlapping $2\times2$ blocks and replace all four tokens in each block with their mean.
Under this proxy, the LPIPS deviation from the paired baseline remains below $0.26$ throughout over 10-second rollout (blue curve in Fig.~\ref{fig:motivation}), suggesting that mixed-granularity patchification layers are learnable within joint denoising windows.

\paragraph{Probing attention-context granularity.}
For selected noise levels, we apply multi-scale self-attention (MSSA, Sec.~\ref{sec:hierarchical_patching}) by spatially subsampling the visible \emph{keys} and \emph{values} based on the stride assigned to each noise level.
This reduces the spatial density of KV-context for high-noise \emph{queries}.
The LPIPS remains below $0.15$ throughout the rollout (red curve in Fig.~\ref{fig:motivation}), showing stable rollout behavior under the resulting mixed-granularity self-attention of joint denoising window.

\section{Methodology}
\subsection{Multi-Scale Patchification and Self-Attention}
\label{sec:hierarchical_patching}
\begin{figure*}[t]
    \centering
    \includegraphics[width=\textwidth]{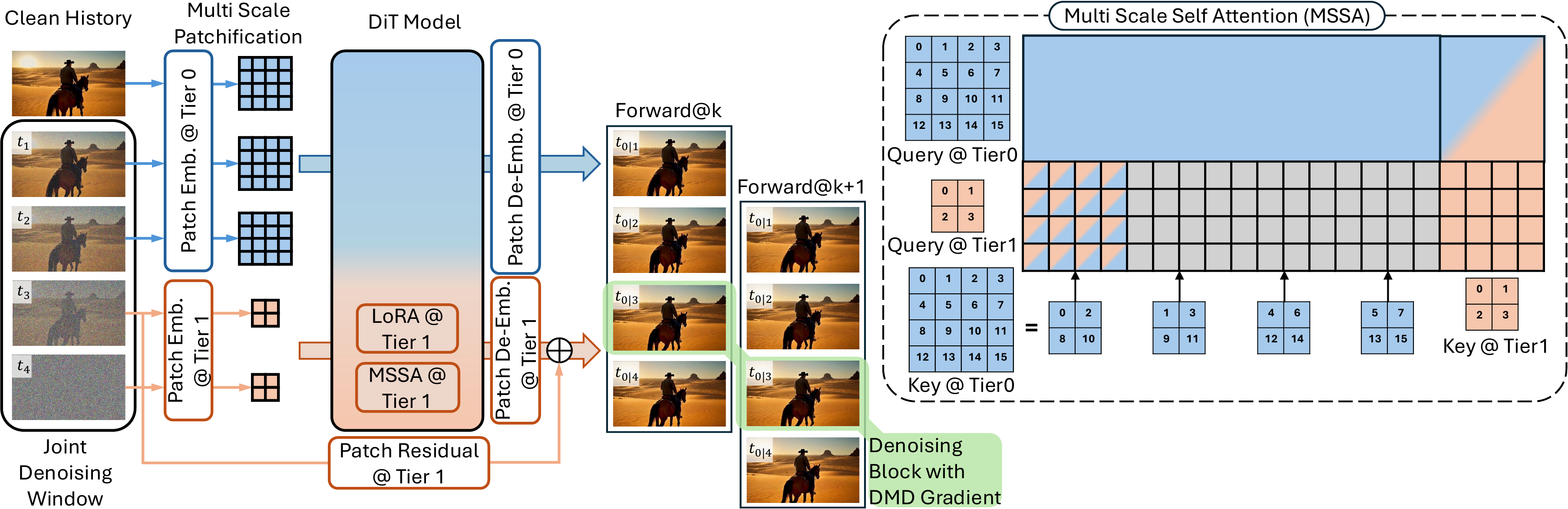}
    \caption{\textbf{Overview of Ms.~Forcing.} Multi-Scale Patchification (MSP) maps the mixed noise levels in a joint denoising window to static patch-size tiers, which are processed by a shared DiT with lightweight tier-specific modules. Multi-Scale Self-Attention (MSSA) matches the spatial density of attention-visible keys and values to each query tier without changing the attention mask. As the window rolls, each state is re-tokenized from coarse to fine; predicted clean states at a shared noise level form the homogeneous sequence used for DMD training.}
    \label{fig:hp_hsa}
\end{figure*}

In this section, we introduce the \emph{Multi-Scale Patchification} (MSP) and \emph{Multi-Scale Self-Attention} (MSSA), which are built upon the RF joint denoising schema.
As illustrated in Fig.~\ref{fig:hp_hsa}, MSP assigns a noise-matched mixed-granularity patch setting within the window, while MSSA matches the spatial density of its attention-visible keys and values to that granularity.
The resulting scale tiers are processed jointly by a shared DiT with lightweight tier-specific modules, without dynamic routing.

\paragraph{Multi-Scale Patchification.}
In the joint denoising window at \emph{forward}$@k$, latent frame $\boldsymbol{x}^{k+l-1}_{t_l}$ at position $l$ is assigned a patch size $p_l$ under the monotonic schedule $p_1\le\cdots\le p_T$, $p_l\in \{p,2p,4p,\cdots\}$, as shown in Fig.~\ref{fig:hp_hsa}.
Each patch size defines a tier with a complete scale-specific processing module, comprising patch embedding and de-embedding layers, LoRA adapters, and a latent residual branch.
The native $p$-tier directly reuses the pretrained model's original pathway and parameters, whereas each coarser tier has its own separately parameterized module; the underlying DiT backbone remains shared across tiers.
To quantify the saving, we compare against RF, which tokenizes all $T$ window states at the native patch size and therefore uses $N_{\mathrm{RF}}=THW/p^2$ tokens per window. 
For a latent frame with spatial size $H\times W$, MSP instead uses
\begin{equation}
    N_l^{\mathrm{MSP}}=\frac{HW}{p_l^2},\qquad
    N_{\mathrm{MSP}}=\sum_{l=1}^{T}N_l^{\mathrm{MSP}},\qquad
    \rho_{\mathrm{tok}}\triangleq\frac{N_{\mathrm{MSP}}}{N_{\mathrm{RF}}}
    =\frac{1}{T}\sum_{l=1}^{T}\left(\frac{p}{p_l}\right)^2.
\label{eq:hp_tokens}
\end{equation}
Here, $\rho_{\mathrm{tok}}$ is the fraction of tokens retained by MSP in the joint denoising window relative to RF.
For the 5-step RF with $(p,p,2p,2p,2p)$ used in our experiments, $\rho_{\mathrm{tok}}=0.55$, corresponding to a $45\%$ token reduction in the joint denoising window.

\paragraph{Multi-Scale Self-Attention.}
Motivated by the attention-context granularity probe in Sec.~\ref{sec:motivation}, MSSA removes the remaining attention redundancy by making each visible non-sink KV context no denser than the query representation, while leaving the attention-sink KVs unchanged (Fig.~\ref{fig:hp_hsa}).
After MSP, let $\mathcal{C}_{\mathrm{sink}}$ and $\mathcal{C}_b$ denote the KV contexts from the attention sink and the non-sink frame at position $b$, respectively, and let $\mathcal{V}(a)$ be the non-sink frame positions visible to queries at position $a$ under the original attention mask.
For the query tokens at position $a$, MSSA constructs the attended KV context as
\begin{equation}
    \mathcal{C}_{a}^{\mathrm{MSSA}}
    =\Big[\,\mathcal{C}_{\mathrm{sink}}\ \Big\|\!
    \operatorname*{Concat}_{b\in\mathcal{V}(a)}
    \operatorname{Subsample}\!\left(
    \mathcal{C}_b,\
    \operatorname{stride}=\max\!\left(1,\frac{p_a}{p_b}\right)
    \right)\Big],
\label{eq:hsa}
\end{equation}
To quantify the incremental gain over MSP, define the normalized token fraction $\eta_l=(p/p_l)^2$ and the two-dimensional spatial stride $s_{ab}=\max(1,p_a/p_b)$.
The resulting reduction in dot-product attention computation is
\begin{equation}
    r_{\mathrm{MSSA}}
    =1-
    \Biggl[\displaystyle\sum_{a=1}^{T}\eta_a
    \Bigl(1+\displaystyle\sum_{b\in\mathcal{V}(a)}\eta_b/s_{ab}^{2}\Bigr)
    \Big/\displaystyle\sum_{a=1}^{T}\eta_a
    \Bigl(1+\displaystyle\sum_{b\in\mathcal{V}(a)}\eta_b\Bigr)\Biggr].
\label{eq:mssa_compute}
\end{equation}
For the 5-step RF layout with one sink frame, one recent-history frame, and five frames in the joint denoising window, substituting the MSP schedule (p,p,2p,2p,2p) into Eq.~\ref{eq:mssa_compute} yields $r_{\mathrm{MSSA}}\approx 12.9\%$.

\subsection{Homogeneous-noise-level assembly Distribution Matching Distillation}
\label{sec:homogeneous_dmd}
Few-step streaming generators are commonly trained with distribution matching distillation (DMD; \citealp{yin2024one,yin2024improved}) on self-generated rollouts~\citep{huang2026self,liu2026rolling} to minimize the reverse KL divergence between real $p_{\text{real}}(\boldsymbol{x})$ and generated distributions $p_{\text{fake}}(\boldsymbol{x})$ across timesteps while reducing the training-inference gap in autoregressive:
\begin{equation}
  \begin{split}
      \nabla_{\theta}\mathcal{L}_{\text{DMD}}&\triangleq \mathbb{E}_{t}(\nabla_{\theta} \mathbb{D}_{\text{KL}}(p_{\text{fake},t}(\boldsymbol{x}_{t})||p_{\text{real},t}(\boldsymbol{x}_{t})))\\
      & \approx -\mathbb{E}_{t}\Big(\int  (s_{\text{real}}(\Psi(G_{\theta}(\epsilon),t),t) -s_{\text{fake}}(\Psi(G_{\theta}(\epsilon),t),t))\frac{\text{d}G_{\theta}(\epsilon)}{\text{d}\theta}\text{d}\epsilon\Big),
  \end{split}
  \label{eq:dmd}
\end{equation}
where $\epsilon \sim \mathcal{N}(0,\mathbf{I})$, $\Psi$ denotes forward diffusion at timestep $t$. In diffusion models, the score function is defined as:
$s_{\text{real}}(\boldsymbol{x}_{t},t)=\nabla_{\boldsymbol{x}_{t}}\log p_{\text{real},t}(\boldsymbol{x}_{t})=-\frac{\boldsymbol{x}_{t}-\alpha_{t}\mu_{\text{real}}(\boldsymbol{x}_{t},t)}{\sigma_{t}^{2}}$,
where $\mu_{\text{real}}$ is the denoised estimate, and $\alpha_{t}, \sigma_{t}$ are noise schedule parameters~\citep{ho2020denoising,nichol2021improved,karras2022elucidating}. DMD freezes pretrained $\mu_{\text{real}}$ (teacher) and updates $\mu_{\text{fake}}$ on generator outputs.

Rolling Forcing sample a subset of non-overlapping windows to construct the predicted clean video for DMD training where gradient computation is performed only on these selected windows, significantly reducing memory usage while retaining effective supervision.
\begin{equation}
    \hat{\boldsymbol{x}}_{0,\mathrm{RF}}^{1:N}
    =\operatorname*{Concat}_{\substack{1\leq i\leq N\\ i\equiv r\ (\mathrm{mod}\ T)}}
    G_{\theta}\!\left(
    \boldsymbol{x}^{i:i+T-1}_{t_{1:T}},t_{1:T},\boldsymbol{x}^{<i}_{0}
    \right),
    \qquad r\sim\mathrm{Uniform}\{0,\ldots,T-1\}.
\label{eq:rf_eq}
\end{equation}
However, Eq.~\ref{eq:rf_eq} concatenates clean predictions from different rollout noise levels $t_{1:T}$, producing a heterogeneous fake sequence with level-dependent temporal marginals. RF mitigates this issue by equally alternating Self-Forcing and Rolling-Forcing updates, but the latter still exposes DMD to variations in clarity and motion that may induce unnatural appearance changes or camera movement.

Instead, at each iteration, we propose H-DMD and sample a window position $j$ and collect the corresponding clean prediction from every consecutive rolling window:
\begin{equation}
    \hat{\boldsymbol{x}}_{0,\mathrm{Our}}^{1:N}
    =\operatorname*{Concat}_{i=1}^{N}
    \pi_j\!\left(
    G_{\theta}\!\left(
    \boldsymbol{x}^{i:i+T-1}_{t_{1:T}},t_{1:T},\boldsymbol{x}^{<i}_{0}
    \right)
    \right),
    \qquad j\sim\mathrm{Uniform}\{1,\ldots,T\}.
\label{eq:homo}
\end{equation}
Here $\pi_j$ selects the clean prediction associated with noise level $t_j$ in each window.
Consequently, every assembled clean predicted video is homogeneous in its source noise level.
By reducing token and attention computation, MSP and MSSA keep training tractable despite backpropagating the DMD gradient through all overlapping joint-denoising windows. 
The complete training procedure is summarized in Algorithm~\ref{alg:training}.


\begin{algorithm}[t]
  \caption{Ms.~Forcing Training with Homogeneous-noise-level DMD}
  \small
  \begin{algorithmic}[1]
    \Require Rollout levels $\{\tau_{0},\tau_{1},\dots,\tau_{T}\}$; patch sizes $p_{1}\le\cdots\le p_{T}$; sequence length $N$
    \Require Generator $G_{\theta}$; fake-score network $s_{\phi}$; frozen teacher score $s_{\mathrm{teacher}}$
    \Loop
      \State Initialize $\mathbf{X}_{\theta}\gets[\,]$, $\mathbf{KV}\gets[\,]$, and warm up the first $T-1$ rolling states
      \State Sample $j\sim\mathrm{Uniform}\{1,\dots,T\}$ \Comment{shared rollout level}
      \For{$i=1,\dots,N$}
        \State Append $\boldsymbol{x}^{i+T-1}_{\tau_T}\sim\mathcal{N}(0,\mathbf{I})$ to form $\boldsymbol{x}^{i:i+T-1}_{\tau_{1:T}}$
        \State $\hat{\mathbf{W}}_{0}^{(i)}\gets G_{\theta}(\boldsymbol{x}^{i:i+T-1}_{\tau_{1:T}},\tau_{1:T},\mathbf{KV})$ \Comment{MSP and MSSA}
        \State $\mathbf{X}_{\theta}.\texttt{append}\big(\pi_j(\hat{\mathbf{W}}_{0}^{(i)})\big)$ \Comment{Eq.~\ref{eq:homo}}
        \State Update $\mathbf{KV}$ with $\operatorname{sg}[\pi_1(\hat{\mathbf{W}}_{0}^{(i)})]$
        \State $\boldsymbol{x}^{i+1:i+T-1}_{\tau_{1:T-1}}\gets\Psi\!\left(\operatorname{sg}[\pi_{2:T}(\hat{\mathbf{W}}_{0}^{(i)})],\tau_{1:T-1}\right)$
      \EndFor
      \State Update $\phi$ with the diffusion denoising loss on $\operatorname{sg}[\mathbf{X}_{\theta}]$
      \If{generator-update step}
        \State Sample DMD timestep $t$ and noise $\boldsymbol{\epsilon}$; set $\mathbf{X}_{t}=\alpha_t\mathbf{X}_{\theta}+\sigma_t\boldsymbol{\epsilon}$
        \State Freeze $s_{\phi}$ and update $\theta$ with Eq.~\eqref{eq:dmd} on $\mathbf{X}_{t}$
      \EndIf
    \EndLoop
  \end{algorithmic}
  \label{alg:training}
\end{algorithm}


\section{Experiments}
\subsection{Implementation Details}
\label{sec:impl}
\textbf{Model.}
We implement Ms.~Forcing with Wan2.1-T2V-1.3B~\citep{wan2025wan} as our base model, which generates 16 FPS videos with a resolution of $832\times480$.
Following Self Forcing~\citep{huang2026self} and Rolling Forcing~\citep{liu2026rolling}, we first initialize the base model with causal attention masking on 16k ODE solution pairs sampled from the base model.\footnote{We use the official ODE init checkpoint provided by \href{https://huggingface.co/gdhe17/Self-Forcing/blob/main/checkpoints/ode_init.pt}{Self Forcing}.}
Then we use another 6K ODE solution pairs from Causal Forcing~\citep{zhu2026causal} to warm up the patch embedding, patch de-embedding, patch residual, and LoRA, using AdamW with a batch size of 8 and learning rate $10^{-4}$ for 800 steps.
We set $T=5$ and each chunk in the joint denoising window contains 3 latent frames. 
In DMD stage, the model is trained for 1400 steps with a batch size of 64 and a trained temporal window of 21 latent frames.  
We use the AdamW optimizer for both the generator $G_\theta$ (learning rate $1.5\times10^{-6}$) and the fake score $s_{\text{gen}}$ (learning rate $4.0\times10^{-7}$). The generator is updated every 5 steps of fake score updates.

\begin{figure}[ht]
    \centering
    \includegraphics[width=\linewidth]{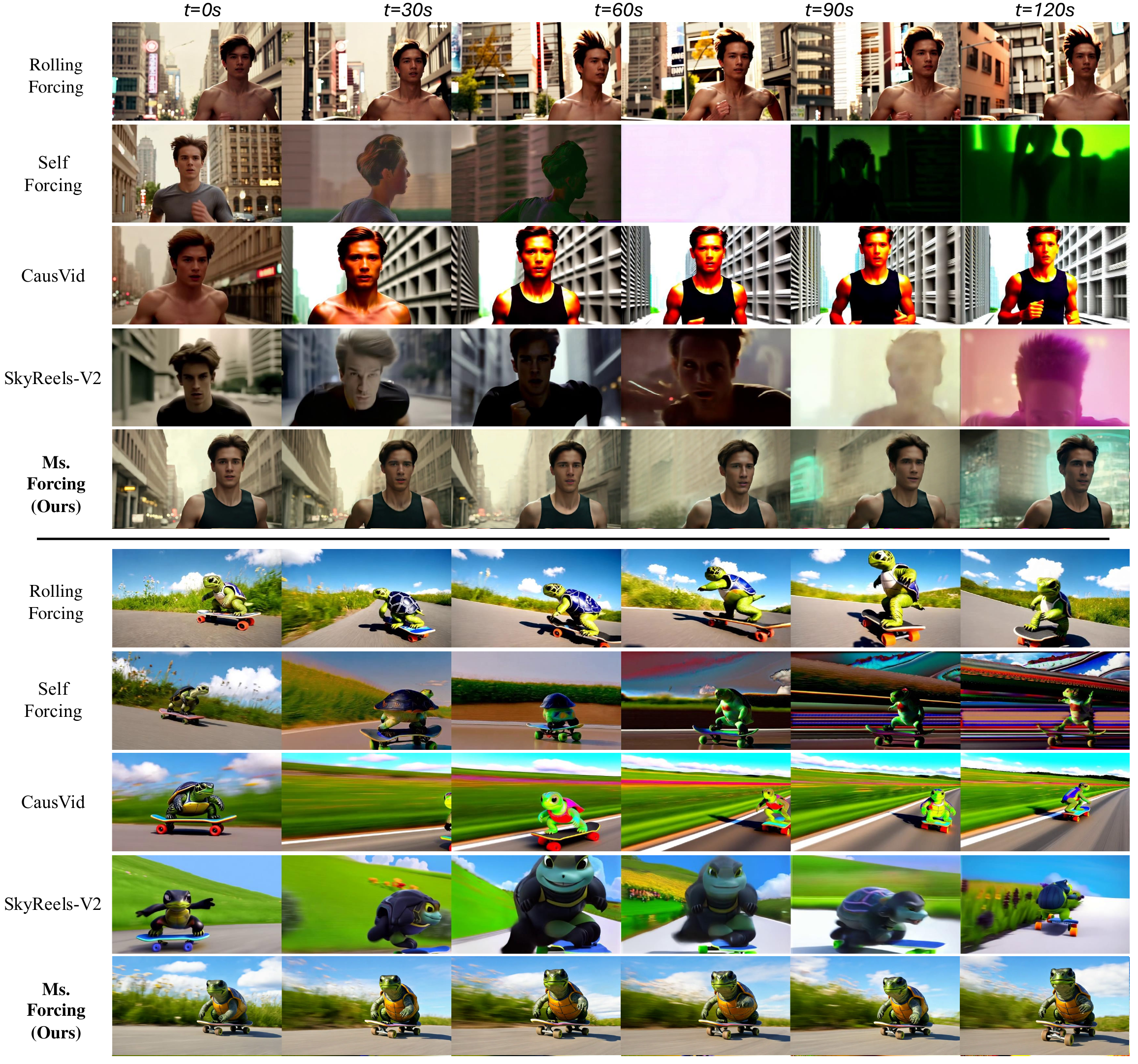}
    \caption{\textbf{Qualitative comparison on long-horizon generation.} Each row shows frames sampled at increasing time from a $30$\,s clip. Compared with the baselines, Ms.~Forcing retains sharpness, consistent color tone, and prompt adherence in late frames, whereas other methods exhibit exposure drift and high-frequency artifacts.}
    \label{fig:qualitative}
\end{figure}

\textbf{Evaluation.} Following Reward Forcing~\citep{lu2026reward}, we evaluate both short-video and long-video generation under the same protocol, excluding its VLM-based scores. For short videos, we generate 5-second clips from the same 946 rewritten official VBench prompts, using five random seeds per prompt, and report the standard VBench total, quality, and semantic scores~\citep{huang2024vbench}. For long videos, we use the first 128 MovieGen benchmark prompts~\citep{polyak2024movie} to generate 60-second videos and evaluate subject consistency, background consistency, motion smoothness, dynamic degree, aesthetic quality, and imaging quality. The individual metrics are normalized and aggregated using the standard VBench coefficients. All videos are generated at 16 FPS and a resolution of $832\times480$. Following the same protocol, we quantify long-video quality drift as the standard deviation of imaging-quality scores over 30 non-overlapping 2-second segments of each video, averaged across the evaluation set; lower values indicate more stable visual quality. We report inference throughput in FPS on a single H200 GPU.

\subsection{Qualitative Results}
Figure~\ref{fig:qualitative} compares Ms.~Forcing with the baselines on long-horizon autoregressive generation lasting over two minutes.
Ms.~Forcing maintains high visual quality and semantic consistency even in later frames, whereas methods with sparser long-range context suffer from exposure drift and high-frequency artifacts.
Additionally, Ms.~Forcing achieves the fastest inference speed among the compared methods while incurring substantially lower training costs than Rolling Forcing.

\subsection{Quantitative Comparisons and Ablations}
\label{sec:ablation}
We compare Ms.~Forcing with representative streaming video generators of comparable scale. These include the Diffusion-Forcing-based SkyReels-V2~\citep{chen2025skyreels,chen2024diffusion} and the distilled causal models CausVid~\citep{yin2025slow}, Self Forcing~\citep{huang2026self}, LongLive~\citep{yang2026longlive}, and Rolling Forcing~\citep{liu2026rolling}. Rolling Forcing is our direct baseline, while the other methods represent complementary strategies for causal distillation and long-horizon generation.

\begin{table}[ht]
  \small
  \centering
  \caption{\textbf{Short-video comparison and cumulative ablation.} H-DMD, MSP, and MSSA are cumulatively added to Rolling Forcing. Best and second-best results are \textbf{bold} and \underline{underlined}, respectively.}
  \begin{tabular}{lccccc}
  \toprule
  \multirow{2}{*}{Model} & \multirow{2}{*}{Params}& \multirow{2}{*}{FPS$\uparrow$} & \multicolumn{3}{c}{VBench evaluation scores $\uparrow$} \\
  \cmidrule(lr){4-6} 
  &&& Total & Quality & Semantic  \\
  \midrule
  SkyReels-V2~\citep{chen2025skyreels} &1.3B& 0.43 & 82.67 & {84.70} & 74.53 \\
  CausVid~\citep{yin2025slow} & 1.3B & 15.92 & 82.88 & 83.93 & 78.69 \\
  Self Forcing~\citep{huang2026self} & 1.3B & 15.92 & \underline{83.80} & \underline{84.59} & 80.64\\
  LongLive~\citep{yang2026longlive}& 1.3B & \underline{19.85} & 83.22 & 83.68 & \textbf{81.37} \\
  \midrule
  \multicolumn{6}{l}{\textit{Component ablation (cumulative)}} \\
  Rolling Forcing~\citep{liu2026rolling}   & 1.3B & 16.35 & 82.88 & 83.63 & 79.88 \\
  \ \ +\,H-DMD                 & 1.3B & 16.35 & 83.10 & 83.57 & \underline{81.22} \\
  \ \ +\,MSP                   & 1.3B & 21.93 & \textbf{83.89} & \textbf{84.61} & 81.01 \\
  \ \ +\,MSSA \textbf{(Ms. Forcing)}  & 1.3B & \textbf{22.84} & 83.36 & 83.92 & 81.13 \\
  \bottomrule
  \end{tabular}
  \label{tab:shortvideo}
\end{table}

\paragraph{Short-video quality and efficiency.}
Table~\ref{tab:shortvideo} reports the VBench results and cumulative component ablation on 5-second videos. 
Compared with baselines, Ms.~Forcing is the fastest, while remaining within $0.44$ of the best total score and $0.24$ of the best semantic score. 
The ablation further disentangles the contribution of each component. 
H-DMD leaves throughput unchanged at $16.35$ FPS but raises the semantic and total scores by $1.34$ and $0.22$, respectively, showing that homogeneous noise-level improves text--video alignment without additional inference cost. 
Adding MSP and MSSA yields significant efficiency gains, \textbf{increasing throughput by $39.6\%$ to $22.84$ FPS while further improving both quality and semantic scores}.
These simultaneous gains in efficiency and generation quality suggest that uniformly processing for different noise level state at the same fine spatial granularity introduces redundancy into the joint denoising window, which may in turn hinder autoregressive generation.

\begin{table*}[ht]
    \small
    \centering
    \caption{\textbf{Long video generation comparison.} H-DMD, MSP, and MSSA are cumulatively added to Rolling Forcing. Best and second-best results are \textbf{bold} and \underline{underlined}, respectively.}
    \resizebox{\textwidth}{!}{%
    \begin{tabular}{lcccccccc}
        \toprule
        \multirow{2}{*}{Model}& \multicolumn{7}{c}{VBench Long Evaluation Scores $\uparrow$} & \multirow{2}{*}{Drift$\downarrow$} \\
        \cmidrule(lr){2-8}
        &Total&Subject&Background&Smoothness&Dynamic&Aesthetic&Imaging Quality& \\
        \midrule
        \cellcolor{lightgray}\textit{Diffusion Forcing}&\cellcolor{lightgray}&\cellcolor{lightgray}&\cellcolor{lightgray}&\cellcolor{lightgray}&\cellcolor{lightgray}&\cellcolor{lightgray}&\cellcolor{lightgray}&\cellcolor{lightgray} \\
        SkyReels-V2~\citep{chen2025skyreels}&75.94&96.43&\underline{96.59}&\textbf{98.91}&39.86&50.76&58.65&7.315\\
        \midrule
        \cellcolor{lightgray}\textit{Distilled Causal}&\cellcolor{lightgray}&\cellcolor{lightgray}&\cellcolor{lightgray}&\cellcolor{lightgray}&\cellcolor{lightgray}&\cellcolor{lightgray}&\cellcolor{lightgray}&\cellcolor{lightgray} \\
        CausVid~\citep{yin2025slow}&77.78&\underline{97.92}&\textbf{96.62}&98.47&27.55&58.39&67.77&2.906\\
        Self Forcing~\citep{huang2026self}&79.34&97.10&96.03&98.48&\textbf{54.94}&54.40&67.61&5.075\\
        LongLive~\citep{yang2026longlive}&79.53&\textbf{97.96}&96.50&\underline{98.79}&35.54&57.81&\underline{69.91}&2.531\\
        \midrule
        \multicolumn{6}{l}{\textit{Component ablation (cumulative)}} \\
        Rolling Forcing~\citep{liu2026rolling} & 79.38 & 97.78 & 96.41 & 98.73 & 33.67 & \underline{60.06} & \textbf{70.66} & \underline{2.227} \\
        \ \ +\,H-DMD & \textbf{80.67} & 97.51 & 95.98 & 98.67 & \underline{51.80} & \textbf{60.11} & 69.72 & 3.041 \\
        \ \ +\,MSP & \underline{79.95} & 97.50 & 96.07 & 98.56 & 48.78 & 59.06 & 68.60 & 2.582 \\
        \ \ +\,MSSA \textbf{(Ms. Forcing)} & 79.66 & 97.90 & 96.58 & 98.70 & 41.85 & 58.89 & 69.04 & \textbf{1.700} \\
        \bottomrule
    \end{tabular}%
    }
    \label{tab:longvideo}
\end{table*}

\paragraph{Long-video quality and drift.}
Table~\ref{tab:longvideo} evaluates 60-second generation, where error accumulation becomes more pronounced.
H-DMD substantially increases the dynamic degree of Rolling Forcing from $33.67$ to $51.80$, indicating stronger motion dynamics over long-horizon rollouts.
However, this improvement comes with increased quality drift.
Removing redundant fine-grained computation from the joint denoising window with MSP and MSSA progressively reduces the drift to $2.582$ and $1.700$, respectively.
This stabilization is accompanied by a gradual decrease in dynamic degree, yet the complete Ms.~Forcing model remains substantially more dynamic than Rolling Forcing while achieving markedly lower drift.
Overall, Ms.~Forcing strikes a favorable balance between motion dynamics and long-horizon stability while maintaining competitive generation quality across the remaining metrics and delivering faster streaming inference speed than Rolling Forcing.

\section{Conclusion}
We presented \textbf{Ms.~Forcing}, an efficient streaming video generation paradigm that reduces the fixed-granularity computation of rolling-window denoising. 
Exploiting the deterministic correspondence between window position and noise level, \emph{Multi-Scale Patchification} (MSP) assigns coarser patches to noisier states and reduces the number of tokens in the joint denoising window by $45\%$. 
\emph{Multi-Scale Self-Attention} (MSSA) further reduces post-MSP dot-product attention computation by $12.9\%$ by matching the density of the visible non-sink KV context to each query's patch granularity while preserving a full-resolution attention sink. 
Both schedules are fixed by window position and therefore remain static and hardware-friendly. 
We also introduced \emph{Homogeneous-Noise-Level DMD} (H-DMD), which replaces RF's heterogeneous fake-video assembly with sequences whose predictions share a common source noise level, reducing the mismatch with inference-time rollouts without auxiliary Self-Forcing training. 
At $832\times480$ resolution, Ms.~Forcing increases throughput from $16.35$ to $22.84$ FPS on a single H200 GPU, a $39.6\%$ improvement over RF. 
It also improves five-second VBench quality and semantic scores by $0.29$ and $1.25$, respectively, and reduces 60-second quality drift from $2.227$ to $1.700$, demonstrating a favorable balance between generation quality, long-horizon stability, and efficiency.

\paragraph{Limitations and future work.}
First, our current experiments focus on the 1.3B-parameter Wan2.1 backbone at a resolution of $832\times480$.
Extending the training and evaluation of Ms.~Forcing to larger models and higher-resolution videos is a natural direction for future work.
Second, the KV-cache update at every rolling step (Line~7 in Alg.~\ref{alg:training}) is necessary for maintaining the autoregressive context, but its current implementation incurs substantial latency.
Consequently, the cache-update overhead partially offsets the acceleration gained by reducing DiT computation with MSP and MSSA.
Optimizing this required update could allow the reduction in DiT computation to translate more fully into end-to-end throughput gains.

\section*{Acknowledgments}
This research was supported by NSF CAREER Award No.~2143576 and a sponsored research award from Meta Inc.

\bibliography{iclr2026_conference}
\bibliographystyle{iclr2026_conference}

\end{document}